\documentclass[runningheads]{llncs}
\usepackage{graphicx}
\usepackage{caption}
\usepackage{url}
\usepackage{amsmath}
\DeclareMathOperator*{\argmin}{\arg\!\min}
\DeclareMathOperator*{\argmax}{\arg\!\max}

\begin{document}
	\title{Explaining Ridesharing: Selection of Explanations for Increasing User Satisfaction}
	%
	\titlerunning{Selection of Explanations for Increasing User Satisfaction}
	%
	\author{David Zar, Noam Hazon, Amos Azaria
		\institute{Computer Science Department, Ariel University, Israel \\
			\email{\{david.zar,noamh,amos.azaria\}@ariel.ac.il}}}
	%
	%
	%
	\maketitle              
	\begin{abstract}
		Transportation services play a crucial part in the development of modern smart cities.  
		In particular, on-demand ridesharing services, which group together passengers with similar itineraries, are already operating in several metropolitan areas.
		These services can be of significant social and environmental benefit, by reducing travel costs, road congestion and $CO_2$ emissions. 
		
		Unfortunately, despite their advantages, not many people opt to use these ridesharing services. We believe that increasing the user satisfaction from the service will cause more people to utilize it, which, in turn, will improve the quality of the service, such as the waiting time, cost, travel time, and service availability. One possible way for increasing user satisfaction is by providing appropriate explanations comparing the alternative modes of transportation, such as a private taxi ride and public transportation. For example, a passenger may be more satisfied from a shared-ride if she is told that a private taxi ride would have cost her $50\%$ more. Therefore, the problem is to develop an agent that provides explanations that will increase the user satisfaction.
		
		We model our environment as a signaling game and show that a rational agent, which follows the perfect Bayesian equilibrium, must reveal all of the information regarding the possible alternatives to the passenger.
		In addition, we develop a machine learning based agent that, when given a shared-ride along with its possible alternatives, selects the explanations that are most likely to increase user satisfaction.
		Using feedback from humans we show that our machine learning based agent outperforms the rational agent and an agent that randomly chooses explanations, in terms of user satisfaction.
	\end{abstract}

	\section{Introduction}
	More than $55\%$ of the world’s population are currently living in urban areas, a proportion that is expected to increase up to $68\%$ by 2050 \cite{united2018}. Sustainable urbanization is a key to successful future development of our society. A key inherent goal of sustainable urbanization is an efficient usage of transportation resources in order to reduce travel costs, avoid congestion, and reduce greenhouse gas emissions. 
	
	
	While traditional services---including buses and taxis---are well established, large potential lies in shared but flexible urban transportation. On-demand ridesharing, where the driver is not a passenger with a specific destination, appears to gain popularity in recent years, and big ride-hailing services such as Uber and Lyft are already offering such services. However, despite the popularity of Uber and Lyft \cite{likeUber2018}, their ridesharing services, which group together multiple passengers (Uber-Pool and Lyft-Line), suffer of low usage \cite{motherboard2016,resons2017}.
	
	
	In this paper we propose to increase the user satisfaction from a given shared-ride, in order to encourage her to use the service more often. That is, we attempt to use a form of persuasive technology \cite{fogg2002persuasive}, not in order to convince users to take a shared ride, but to make them feel better with the choice they have already made, and thus improve their attitude towards ride-sharing.
	It is well-known that one of the most influencing factors for driving people to utilize a specific service is to increase their satisfaction form the service (see for example, \cite{singh2006importance}). Moreover, if people will be satisfied and use the the service more often it will improve the quality of the service, such as the waiting time, cost, travel time, and service availability, which in turn further increase the user satisfaction.

	One possible way for increasing user satisfaction is by providing appropriate explanations \cite{bradley2009dealing}, during the shared ride or immediately after the passenger has completed it. Indeed, in recent years there is a growing body of literature that deals with explaining decisions made by AI systems \cite{gunning2017explainable,kleinerman2018providing}.
	In our ridesharing scenario, a typical 
	approach would attempt to explain the entire assignment of all passengers to all vehicles. Clearly, a passenger is not likely to be interested in such an explanation, since she is not interested in the assignment of other passengers to other vehicles.
	A passenger is likely to only be interested with her own current shared-ride when compared to other alternative modes of transportation, such as a private taxi ride or public transportation.
	
	Comparing the shared-ride to other modes of transportation may provide many different possible explanations. 
	For example, consider a shared-ride that takes $20$ minutes and costs $\$10$. The passenger could have taken a private taxi that would have cost $\$20$. Alternatively, the passenger could have used public transportation, and such a ride would have taken $30$ minutes. A passenger is not likely to be aware of the exact costs and riding times of the other alternatives, but she may have some estimations. The agent, on the other hand, has access to many sources of information, and it can thus provide the exact values as explanations. Clearly, the agent is not allowed to provide false information. The challenge is to design an agent that provides the appropriate explanation in any given scenario.
	
	We first model our environment as a signaling game~\cite{spence1974market}, which models the decision of a rational agent whether to provide the exact price (i.e., the cost or the travel time) of a possible alternative mode of transportation, or not.  
	In this game there are three players: nature, the agent and the passenger. 
	Nature begins by randomly choosing a price from a given distribution; this distribution is known both to the agent and the passenger. The agent observes the price and decides whether to disclose this price to the passenger or not. The passenger then determines her current expectation over the price of the alternative. The goal of the agent is to increase the passenger satisfaction, and thus it would like the passenger to believe that the price of the alternative is higher than the price of the shared-ride as much as possible. We use the standard solution concept of Perfect Bayesian Equilibrium (PBE)~\cite{fudenberg1991perfect} and show that a rational agent must reveal all of the information regarding the price of the possible alternative to the passenger.
	
	Interacting with humans and satisfying their expectations is a very complex task. Research into humans' behavior has found that people often deviate from what is thought to be the rational behavior, since they are affected by a variety of (sometimes conflicting) factors: a lack of knowledge of one's own preferences, framing effects, the interplay between emotion and cognition, future discounting, anchoring and many other effects~\cite{tversky81,Loewenstein00,ArielyAnchor,camerer03}.
	Therefore, algorithmic approaches that use a pure theoretically analytic objective often perform poorly with real humans~\cite{Peledetal11,azaria2015strategic,nay2016predicting}.
	We thus develop an Automatic eXplainer for Increasing Satisfaction (AXIS) agent, that when given a shared-ride along with its possible alternatives selects the explanations that are most likely to increase user satisfaction.
	
	For example, consider again the setting in which a shared-ride takes $20$ minutes and costs $\$10$. 
	The passenger could have taken a private taxi that would have taken $15$ minutes, but would have cost $\$20$. Alternatively, the passenger could have used public transportation. Such a ride would have taken $30$ minutes, but would have cost only $\$5$. 
	A \emph{human} passenger may be more satisfied from the shared-ride if she is told that a private taxi would have cost her $100\%$ more. Another reasonable explanation is that a public transportation would have taken her $10$ minutes longer. It may be even better to provide both explanations. However, providing an explanation that public transportation would have cost $50\%$ less than the shared-ride is less likely to increase her satisfaction. Indeed, finding the most appropriate explanation depends on the specific parameters of the scenario. For example, if public transportation still costs $\$5$ but the shared ride costs only $\$6$, providing an explanation that public transportation would have cost only $\$1$ less than the shared-ride may now become an appropriate explanation. 
	
	For developing the AXIS agent we utilize the following approach.
	We collect data from human subjects on which explanations they believe are most suitable for different scenarios. AXIS then uses a neural network to generalize this data in order to provide appropriate explanations for any given scenario. Using feedback from humans we show that AXIS outperforms the PBE agent and an agent that randomly chooses explanations. That is, human subjects that were faced with shared-ride scenarios, were more satisfied from the ride given the explanations selected by AXIS, than by the same ride when shown all explanations and when the explanations were randomly selected.
	
	The contributions of this paper are threefold:
	\begin{itemize}
		\item The paper introduces the problem of automatic selection of explanations in the ridesharing domain, for increasing user satisfaction. The set of explanations consists of alternative modes of transportation.
		\item We model the explanation selection problem as a signaling game and determine the unique set of Perfect Bayesian Equilibria (PBE).
		\item We develop the AXIS agent, which learns from how people choose appropriate explanations, and show that it outperforms the PBE agent an agent that randomly chooses explanations, in terms of user satisfaction. 
	\end{itemize}

	\section{Related Work}

	Most work on ridesharing has focused on the assignment of passengers to vehicles. See the comprehensive surveys by Parragh et al.~\cite{parragh2008a,parragh2008b}, and a recent survey by Psaraftis et al.~\cite{psaraftis2016dynamic}.
	In particular, the dial-a-ride problem (DARP) is traditionally distinguished from other problems of ridesharing since transportation cost and user inconvenience must be weighed against each other in order to provide an appropriate solution~\cite{cordeau2003tabu}. Therefore, the DARP typically includes more quality constraints that aim at capturing the user's inconvenience. We refer to a recent survey on DARP by Molenbruch et al.~\cite{molenbruch2017}, which also makes this distinction.
	In recent years there is an increasing body of works that concentrate on the passenger's satisfaction during the assignment of passengers to vehicles \cite{lin2012research,levinger2020human,schleibaum2020human}.
	Similar to these works we are interested in the satisfaction of the passenger, but instead of developing assignment algorithms 
	(e.g., \cite{bistaffa2017cooperative}), we emphasize the importance of explanations of a given assignment.
	
	A domain closely related to ridesharing is car-pooling. In this domain, ordinary drivers, may opt to take an additional passenger on their way to a shared destination. The common setting of car-pooling is within a long-term commitment between people to travel together for a particular purpose, where ridesharing is focused on single, non-recurring trips. 
	Indeed, several works investigated car-pooling that can be established on a short-notice, and they refer to this problem as ridesharing~\cite{agatz2012optimization}. 
	In this paper we focus on ridesharing since it seems that our explanations regarding the alternative modes of transportation are more suitable for this domain (even though they might be also helpful for car-pooling).

	In our work we build an agent that attempts to influence the attitude of the user towards ridesharing. Our agent is thus a form of persuasive technology \cite{oinas2008systematic}. 
	Persuasion of humans by computers or technology has raised great interest in the literature. In his book \cite{fogg2002persuasive}, Fogg surveyed many technologies to be successful. One example of such a persuasion technology (pg. 50) is bicycle connected to a TV; as one pedals at a higher rate, the image on the TV becomes clearer, encouraging humans to exercise at higher rates. Another example is the Banana-Rama slot machine, which has characters that celebrate every time the gambler wins. 
	Overall, Fogg describes $40$ persuasive strategies. Other social scientists proposed various classes of persuasive strategies: Kellermann and Tim provided over $100$ groups~\cite{kellermann1994classifying}, and Cialdini proposed six principles of influence \cite{cialdini2001harnessing}.
	More specifically, Anagnostopoulou et al. \cite{anagnostopoulou2018persuasive} survey persuasive technologies for sustainable mobility, some of which consider ridesharing. The methods mentioned by Anagnostopoulou et al. include several persuasive strategies such as self-monitoring, challenges \& goal setting, social comparison, gamification, tailoring, suggestions and rewards.
	Overall, unlike most of the works on persuasive technology, our approach is to selectively provide information regarding alternative options. This information aims at increasing the user satisfaction from her action, in order to change her attitude towards the service.

	
	There are other works in which an agent provides information to a human user (in the context of the roads network) for different purposes. For example, Azaria et al.~\cite{azaria2012giving,azaria2012strategic,azaria2015strategic} develop agents that provide information or advice to a human user in order to convince her to take a certain route. Bilgic and Mooney \cite{bilgic2005explaining} present methods for explaining the decisions of a recommendation system to increase the user satisfaction. In their context, user satisfaction is interpreted only as an accurate estimation of the item quality.

	
	Explainable AI (XAI) is another domain related to our work \cite{core2006building,gunning2017explainable,carvalho2019machine}. In a typical XAI setting, the goal is to explain the output of the AI system to a human. This explanation is important for allowing the human to trust the system, better understand, and to allow transparency of the system's output \cite{adadi2018peeking}. Other XAI systems are designed to provide explanations, comprehensible by humans, for legal or ethical reasons \cite{doran2017does}. For example, an AI system for the medical domain might be required to explain its choice for recommending the prescription of a specific drug \cite{holzinger2017we}. 
	Despite the fact that our agent is required to provide explanations to a human, our work does not belong to the XAI settings. In our work the explanations do not attempt to explain the output of the system to a passenger but to provide additional information that is likely to increase the user's satisfaction from the system. Therefore, our work can be seen as one of the first instances of x-MASE \cite{Kraus2019ai}, explainable systems for multi-agent environments.

	\section{The PBE Agent}
	
	We model our setting with the following signaling game.
	We assume that there is a given random variable $X$ with a prior probability distribution over the possible prices of a given alternative mode of transportation. The possible values of $X$ are bounded within the range $[min,max]$\footnote{Without loss of generality, we assume that $Pr(X=min)>0$ for a discrete distribution, and $F_X(min+\epsilon)>0$ for a continuous distribution, for every $\epsilon > 0$.}. 
	
	The game is composed of three players: nature, player 1 (agent) and player 2 (passenger). It is assumed that both players are familiar with the prior distribution over $X$.
	Nature randomly chooses a number $x$ according to the distribution over $X$.
	The agent observes the number $x$ and her possible action, denoted $a_1$, is either  $\varphi$ (quiet) or $x$ (say). 
	That is, we assume that the agent may not provide false information. This is a reasonable assumption, since providing false information is usually prohibited by the law, or may harm the agent's reputation.
	The passenger observes the agent's action and her action, denoted $a_2$, is any number in the range $[min,max]$.
	The passenger's action essentially means setting her estimate about the price of the alternative.
	In our setting the agent would like the passenger to think that the price of the alternative is as high as possible, while the passenger would like to know the real price. Therefore, we set the utility for the agent to $a_2$ and the utility of the passenger to $-(a_2-x)^2$. Note that we did not define the utility of the passenger to be simply $-|a_2-x|$, since we want the utility to highly penalize a large deviation from the true value.
	
	We first note that if the agent plays $a_1 \neq \varphi$ then the passenger knows that $a_1$ is nature's choice. Thus, a rational passenger would play $a_2=a_1$. On the other hand, if the agent plays $a_1=\varphi$ then the passenger would have some belief about the real price, which can be the original distribution of nature, or any other distribution. We show that the passenger's best response is to play the expectation of this belief. Formally, 
	\begin{lemma}
		\label{lemma:belief}
		Assume that the agent plays $a_1=\varphi$, and let $Y$ be a belief over $x$. That is, $Y$ is a random variable with a distribution over $[min,max]$. Then, $\argmax_{a_2} E[-(a_2-Y)^2] = E[Y]$. \end{lemma}
	\begin{proof}
		Instead of maximizing $E[-(a_2-Y)^2]$ we can minimize $E[(a_2-Y)^2]$. In addition, $E[(a_2-Y)^2] = E[(a_2)^2] -2E[a_2 Y] + E[Y^2] = (a_2)^2 -2a_2 E[Y] + E[Y^2]$. By differentiating we get that
		\[ \frac d {da_2} \left((a_2)^2 -2a_2 E[Y] + E[Y^2]\right) = 2a_2 -2E[Y].\]
		The derivative is $0$ when $a_2 = E[Y]$ and the second derivative is positive; this entails that
		\[\argmin_{a_2} \left((a_2)^2 -2a_2 E[Y] + E[Y^2]\right) = E[Y]\]
		\qed \end{proof}
	
	Now, informally, if nature chooses a ``high'' value of $x$, the agent would like to disclose this value by playing $a_1=x$. One may think that if nature chooses a ``low'' value of $x$, the agent would like to hide this value by playing $a_1=\varphi$. However, since the user adjusts her belief accordingly, she will play $E[X|a_1=\varphi]$. Therefore, it would be more beneficial for the agent to reveal also low values that are greater than $E[X|a_1=\varphi]$, which, in turn, will further reduce the new $E[X|a_1=\varphi]$. Indeed, Theorem~\ref{thm:pbe} shows that a rational agent should always disclose the true value of $x$, unless $x=min$. If $x=min$ the agent can play any action, i.e., $\varphi$, $min$ or any mixture of $\varphi$ and $min$. 
	We begin by applying the definition of PBE to our signaling game.   
	\begin{definition}
		A tuple of strategies and a belief, $(\sigma_1, \sigma_2, \mu_2)$, is said to be a perfect Bayesian equilibrium in our setting if the following hold:
		\begin{enumerate}
			\item The strategy of player 1 is a best response strategy. That is, given $\sigma_2$ and $x$, deviating from $\sigma_1$ does not increase player 1's utility. 
			\item The strategy of player 2 is a best response strategy. That is, given $a_1$, deviating from $\sigma_2$ does not increase player 2's expected utility according to her belief.
			\item $\mu_2$ is a consistent belief.
			That is, $\mu_2$ is a distribution over $x$ given $a_1$, which is consistent with $\sigma_1$ (following Bayes rule, where appropriate).
		\end{enumerate}
	\end{definition}
	
	\begin{theorem}
		\label{thm:pbe}
		A tuple of strategies and a belief, $(\sigma_1, \sigma_2, \mu_2)$, is a PBE if and only if:
		\begin{itemize}
			\item $\sigma_1(x)=\begin{cases} x: & x>min \\ \text{anything}: & x=min \end{cases}$
			\item $\sigma_2(a_1)= \begin{cases} a_1: & a_1 \neq \varphi \\ min: & a_1=\varphi \end{cases}$
			\item $\mu_2(x=a_1| a_1\neq \varphi) = 1$ and $\mu_2(x=min| a_1=\varphi)=1$.
		\end{itemize}
	\end{theorem}
	\begin{proof}
		($\Leftarrow$) Such a tuple is a PBE: $\sigma_1$ is a best response strategy, since the utility of player 1 is $x$ if $a_1=x$ and $min$ if $a_1=\varphi$. Thus, playing $a_1=x$ is a weakly dominating strategy. 
		$\sigma_2$ is a best response strategy, since it is the expected value of the belief $\mu_2$, and thus it is a best response  according to Lemma \ref{lemma:belief}.
		Finally, $\mu_2$ is consistent: 
		If $a_1=\varphi$ and according to $\sigma_1$ player 1 plays $\varphi$ with some probability (greater than 0), then according to Bayes rule $\mu_2 (x=min| a_1=\varphi)=1$. Otherwise, Bayes rule cannot be applied (and it is thus not required).
		If $a_1\neq \varphi$, then by definition $x=a_1$, and thus $\mu_2(x=a_1| a_1\neq \varphi)=1$.
		
		($\Rightarrow$)
		Let $(\sigma_1, \sigma_2, \mu_2)$ be a PBE.
		It holds  that $\mu_2(x=a_1| a_1\neq \varphi) = 1$ by Bayes rule, implying that if $a_1\neq\varphi$, $\sigma_2(a_1)=a_1$. Therefore, when $a_1=x$ the utility of player 1 is $x$. 
		
		We now show that $\sigma_2(a_1=\varphi) = min$.
		Assume by contradiction that $\sigma_2(a_1=\varphi)\neq min$ (or $p(\sigma_2(a_1=\varphi)=min) < 1$), then $E[\sigma_2(\varphi)]=c>min$.
		We now imply the strategy of player 1. There are three possible cases: if $x>c$, then $a_1=x$ is a strictly dominating strategy. If $x < c$, then $a_1=\varphi$ is a strictly dominating strategy.
		If $x=c$, there is no advantage for either playing $\varphi$ or $x$; both options give player 1 a utility of $c$, and thus she may use any strategy, i.e.:
		$\sigma_1(x)=\begin{cases}x: & x>c \\ \varphi: & x<c \\ \text{anything}: & x=c.\end{cases}$
		
		Given this strategy, we need to apply Bayes rule to derive $\mu_2(x| a_1=\varphi)$. By $\sigma_1$, it is possible that $a_1=\varphi$ only if $x\leq c$. That is, $\mu_2(x>c| a_1=\varphi)=0$ and $\mu_2(x \leq c| a_1=\varphi)=1$. Therefore, the expected value of the belief, $c^\prime = E[\mu_2(x|a_1=\varphi)]$, and according to Lemma \ref{lemma:belief}, $\sigma_2(\varphi) = c^\prime$. However, $c^\prime = E[\mu_2(x|a_1=\varphi)] \leq E[x | x\leq c]$ which is less than $c$, since $c>min$. That is,
		$E[\sigma_2(\varphi)]=c^\prime < c$, which is a contradiction.
		%
		Therefore, the strategy for player 2 in every PBE is determined. In addition, since $\sigma_2(\varphi) = E[\mu_2(x| a_1=\varphi)]$ according to Lemma~\ref{lemma:belief}, then $\mu_2(x| a_1=\varphi)=min$, and the belief of player 2 in every PBE is also determined.
		
		We end the proof by showing that for $x>min$, $\sigma_1(x)=x$. Since $\sigma_2$ is determined, the utility of player 1 is $min$ if $a_1=\varphi$ and $x$ if $a_1=x$. Therefore, when $x>min$, playing $a_1=x$ is a strictly dominating strategy.
		
		
		\qed \end{proof}
	
	The provided analysis can be applied to any alternative mode of transportation and to any type of price (e.g. travel-time or cost). We thus conclude that the PBE agent must provide all of the possible explanations.

	%
	%
	
	\section{The AXIS Agent}
	\label{sec:AXIS}
	The analysis in the previous section is theoretical in nature. However, several studies have shown that algorithmic approaches that use a pure theoretically analytic objective often perform poorly with real humans. Indeed, we conjecture that an agent that selects a subset of explanations for a given scenario will perform better than the PBE agent. 
	In this section we introduce our Automatic eXplainer for Increasing Satisfaction (AXIS) agent. 
	The AXIS agent has a set of possible explanations, and the agent needs to choose the most appropriate explanations for each scenario. Note that we do not limit the number of explanations to present for each scenario, and thus AXIS needs also to choose how many explanations to present. AXIS was built in $3$ stages.
	
	First, an initial set of possible explanations needs to be defined. We thus consider the following possible classes of factors of an explanation. Each explanation is a combination of one factor from each class:
	\begin{enumerate}
		\item Mode of alternative transportation: a private taxi ride or public transportation.
		\item Comparison criterion: time or cost.
		\item Visualization of the difference: absolute or relative difference.
		\item Anchoring: the shared ride or the alternative mode of transportation perspective.
	\end{enumerate}
	For example, a possible explanation would consist of a private taxi for class $1$, cost for class $2$, relative for class $3$, and an alternative mode of transportation perspective for class $4$. That is, the explanation would be ``a private taxi would have cost $50\%$ more than a shared ride''. Another possible explanation would consist of public transportation for class $1$, time for class $2$, absolute for class $3$, and a shared ride perspective for class $4$. That is, the explanation would be ``the shared ride saved $10$ minutes over  public transportation''. Overall, there are $2^4 = 16$ possible combinations. In addition, we added an explanation regarding the saving of $CO_2$ emission of the shared ride, so there will be an alternative explanation for the case where the other options are not reasonable.
	Note that the first two classes determine which information is given to the passenger, while the later two classes determine how the information is presented. We denote each possible combination of choosing form the first two classes as a \textit{information setting}. We denote each possible combination of choosing form the latter two classes as a \textit{presentation setting}.
	%
	
	Presenting all $17$ possible explanations with the additional option of ``none of the above'' requires a lot of effort from the human subjects to choose the most appropriate option for each scenario. Thus, in the second stage we collected data from human subjects regarding the most appropriate explanations, in order to build a limited subset of explanations. Recall that there are $4$ possible information settings and $4$ possible presentation settings. We selected for each information setting the corresponding presentation setting that was chosen (in total) by the largest number of people. We also selected the second most chosen presentation setting for the information setting that was chosen by the largest number of people. Adding the explanation regarding the $CO_2$ emissions we ended with $6$ possible explanations.
	
	In the final stage we collected again data from people, but we presented only the $6$ explanation to choose from. This data was used by AXIS to learn which explanations are appropriate for each scenario. 
	AXIS receives the following $7$ features as an input: the cost and time of the shared ride, the differences between the cost and time of the shared ride and the alternatives (i.e., the private ride and the public transportation), and the amount of $CO_2$ emission saved when compared to a private ride.
	AXIS uses a neural network with two hidden layers, one with $8$ neurons and the other one with $7$ neurons, and the logistic activation function (implemented using Scikit-learn \cite{scikit-learn}). The number of neurons and hidden layers was determined based on the performance of the network.
	AXIS used $10\%$ of the input as a validation set (used for early stopping) and $40\%$ as the test set. AXIS predicts which explanations were selected by the humans (and which explanations were not selected) for any given scenario. 
	

	\section{Experimental Design}
	In this section we describe the design of our experiments. Since AXIS generates explanations for a given assignment of passengers to vehicles, we need to generate assignments as an input to AXIS. To generate the assignments we first need a data-set of ride requests.
	
	To generate the ride requests we use the New York city taxi trip data-set \footnote{\url{https://data.cityofnewyork.us/Transportation/2016-Green-Taxi-Trip-Data/hvrh-b6nb}}, which was also used by other works that evaluate ridesharing algorithms (see for example, \cite{lin2016model,biswas2017profit}). We use the data-set from 2016, since it contains the exact GPS locations for every ride. 
	
	We note that the data-set contains requests for taxi rides, but it does not contain a data regarding shared-rides. We thus need to generate assignments of passengers to taxis, based on the requests from the data-set. Now, if the assignments are randomly generated, it may be hard to provide reasonable explanations, and thus the evaluation of AXIS in these setting is problematic. We thus concentrate on requests that depart from a single origin but have different destinations, since a brute force algorithm can find the optimal assignment of passengers to taxis in this setting.
	
	We use the following brute force assignment algorithm. The algorithm receives $12$ passengers and outputs the assignment of each passenger to vehicle that minimizes the overall travel distance. We assume that every vehicle can hold up-to four passengers.
	The brute force assignment algorithm recursively considers all options to partition the group of $12$ passengers to subsets of up to four passengers. We note that there are $3,305,017$ such possible partitions. 
	The algorithm then solves the 
	Travel Salesman Problem (TSP) in each group, by exhaustive search, to find the cheapest assignment. 
	Solving the TSP problem on 4 destinations (or less) is possible using exhaustive search since there are only $4!=24$ combinations. The shortest path between each combination is solved using a shortest distance matrix between all locations. 
	In order to compute this matrix we downloaded the graph that represents the area of New York from Open Street Map (using OSMnx \cite{boeing2017osmnx}), and ran the Floyd-Warshall's algorithm. 
	
\begin{figure}[hbpt]
		\centering
		\includegraphics[width=2.7in]{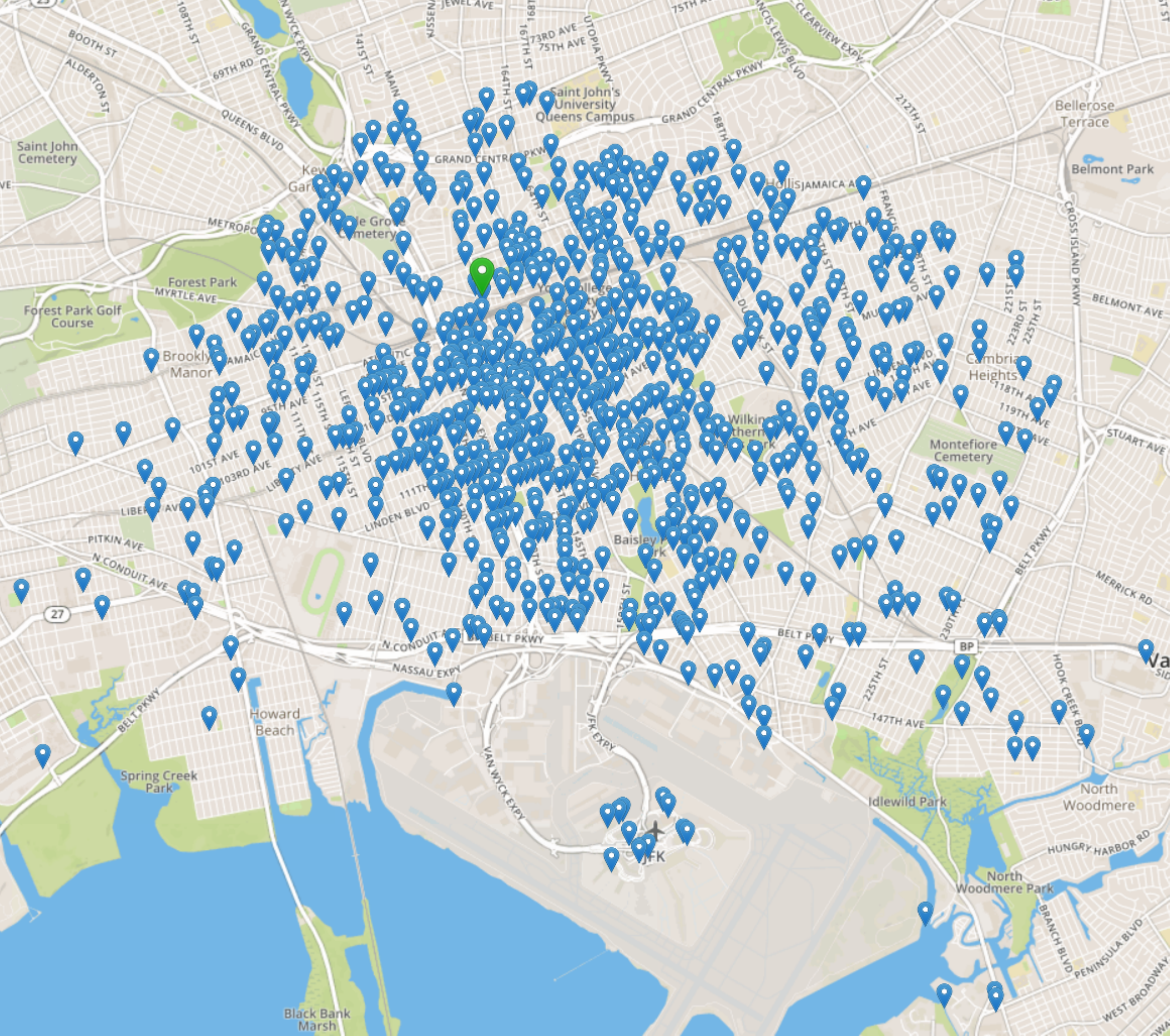} 
		\caption{A map depicting the origin (in green) and destinations (in blue) of all rides considered.}
		\label{fig:destinations}
	\end{figure}

	
	We set the origin location to JFK Station, Sutphin Blvd-Archer Av, and the departing time to 11:00am. See Figure \ref{fig:destinations} where the green location is the origin, and the blue locations are the destinations.
	
	In order to calculate the duration of the rides we use Google Maps (through Google Maps API). Specifically, the duration of the private taxi ride was obtained using ``driving'' mode, and the duration of the public transportation was obtained using ``transit'' mode. 
	The duration of the shared-ride was obtained using ``driving'' mode 
	with the last passenger's destination as the final destination of the ride and the destinations of the other passengers as way-points. The duration for a specific passenger was determined by using the time required to reach her associated way-point.
	
	In order to calculate the cost of the private ride we use Taxi Fare Finder (through their API)\footnote{\url{https://www.taxifarefinder.com/}}. The cost for public transportation was calculated by the number of buses required (as obtained through Google Maps API), multiplied by $\$2.5$ (the bus fare). The cost for the shared-ride was obtained from Taxi Fare Finder. Since this service does not support a ride with way-points, we obtained the cost of multiple taxi rides, but we included the base price only once. Note that this is the total cost of the shared-ride.
	The cost for a specific passenger was determined by the proportional sharing pricing function \cite{fishburn1983fixed}, which works as follows. Let $c_{p_i}$ be the cost of a private ride for passenger $i$, and let $total_s$ be the total cost of the shared ride. In addition, let $f=\frac{total_s}{\sum_i c_{p_i}}$. The cost for each passenger is thus $f \cdot c_{pi}$.
	
	We ran 4 experiments in total. Two experiments were used to compose AXIS (see Section~\ref{sec:AXIS}), and the third and fourth experiments compared the performance of AXIS with that of non-data-driven agents (see below). 
	All experiments used the Mechanical Turk platform, a crowd-sourcing platform that is widely used for running experiments with human subjects \cite{amir2012economic,paolacci2010running}. 
	Unfortunately, since participation is anonymous and linked to monetary incentives, experiments on a crowd-sourcing platform can attract participants who do not fully engage in the requested tasks \cite{turner2012using}. Therefore, the subjects were required to have at least $99\%$ acceptance rate and were required to have previously completed at least $500$ Mechanical Turk Tasks (HITs). In addition, we added an attention check question for each experiment, which can be found in the Appendix. 
	
	In the first two experiments, which were designed for AXIS to learn what people believe are good explanations, the subjects were given several scenarios for a shared ride. The subjects were told that they are representatives of a ride sharing service, and that they need to select a set of explanations that they believe will increase the customer's satisfaction. Each scenario consists of a shared-ride with a given duration and cost. 
	
	In the third experiment we evaluate the performance of AXIS against the PBE agent. 
	The subjects were given $2$ scenarios. Each scenario consists of a shared-ride with a given duration and cost and it also contains either the explanations that are chosen by AXIS or the information that the PBE agent provides: 
	the cost and duration a private ride would take, and the cost and the duration that public transportation would have taken. The subjects were asked to rank their satisfaction from each ride on a scale from 1 to 7.
	
	In the forth experiment we evaluate the performance of AXIS against a random baseline agent.
	The random explanations were chosen as follows: first, a number between $1$ and $4$ was uniformly sampled. This number determined how many explanations will be given by the random agent. This range was chosen since over $93\%$ of the subjects selected between $1$ and $4$ explanations in the second experiment.
	Recall that there are 4 classes of factors that define an explanation, where the fourth class is the anchoring perspective (see Section~\ref{sec:AXIS}). The random agent sampled explanations uniformly, but it did not present two explanations that differ only by their anchoring perspective.
	%
	The subjects were again given $2$ scenarios. Each scenario consists of a shared-ride with a given duration and cost and it also contains either the explanations that are chosen by AXIS or the explanations selected by the random agent. The subjects were asked to rank their satisfaction from each ride.
	The exact wording of the instructions for the experiments can be found in the Appendix. 
	
	$953$ subjects participated in total, all from the USA.
	The number of subjects in each experiment and the number of scenarios appear in Table~\ref{tbl:participants}. 
	Tables \ref{tbl:gender} and \ref{tbl:education} include additional demographic information on the subjects in each of the experiments. The average age of the subjects was $39$. 
	\begin{table}
		\centering
		\begin{tabular}{ c|c c c c c } 
			\hline
			& \#1 & \#2 & \#3 & \#4 & Total\\
			\hline
			Number of subjects & 343 & 180 & 156 & 274 & 953\\
			Scenarios per subject & 2 & 4 & 2 & 2 & -\\
			Total scenarios & 686 & 720 & 312 & 548 & 3266\\
			\hline
		\end{tabular}
		\caption{Number of subjects and scenarios in each of the experiments.}
		\label{tbl:participants}
	\end{table}
	\begin{table}
		\centering
		\begin{tabular}{ c|c c c c c} 
			\hline
			& \#1 & \#2 & \#3 & \#4 & Total\\
			\hline
			Male & 157 & 66 & 52 & 117 & 392\\
			Female & 183 & 109 & 104 & 153 & 549\\
			Other or refused & 3 & 5 & 0 & 4 & 12\\
			\hline
		\end{tabular}
		\caption{Gender distribution for each of the experiments.}
		\label{tbl:gender}
	\end{table}
	\begin{table}
		\centering
		\begin{tabular}{ c|c c c c c } 
			\hline
			& \#1 & \#2 & \#3 & \#4 & Total\\
			\hline
			High-school & 72 & 39 & 38 & 80 & 229\\
			Bachelor & 183 & 86 & 84 & 131 & 484\\
			Master & 60 & 29 & 37 & 46 & 172\\
			PhD & 15 & 2 & 0 & 10 & 27\\
			Trade-school & 8 & 4 & 5 & 10 & 27\\
			Refused or did not respond & 5 & 3 & 0 & 6 & 14\\
			\hline
		\end{tabular}
		\caption{Education level for each of the experiments.}
		\label{tbl:education}
	\end{table}

	\begin{figure}[hbpt]
		\centering
		\includegraphics[width=0.87\textwidth]{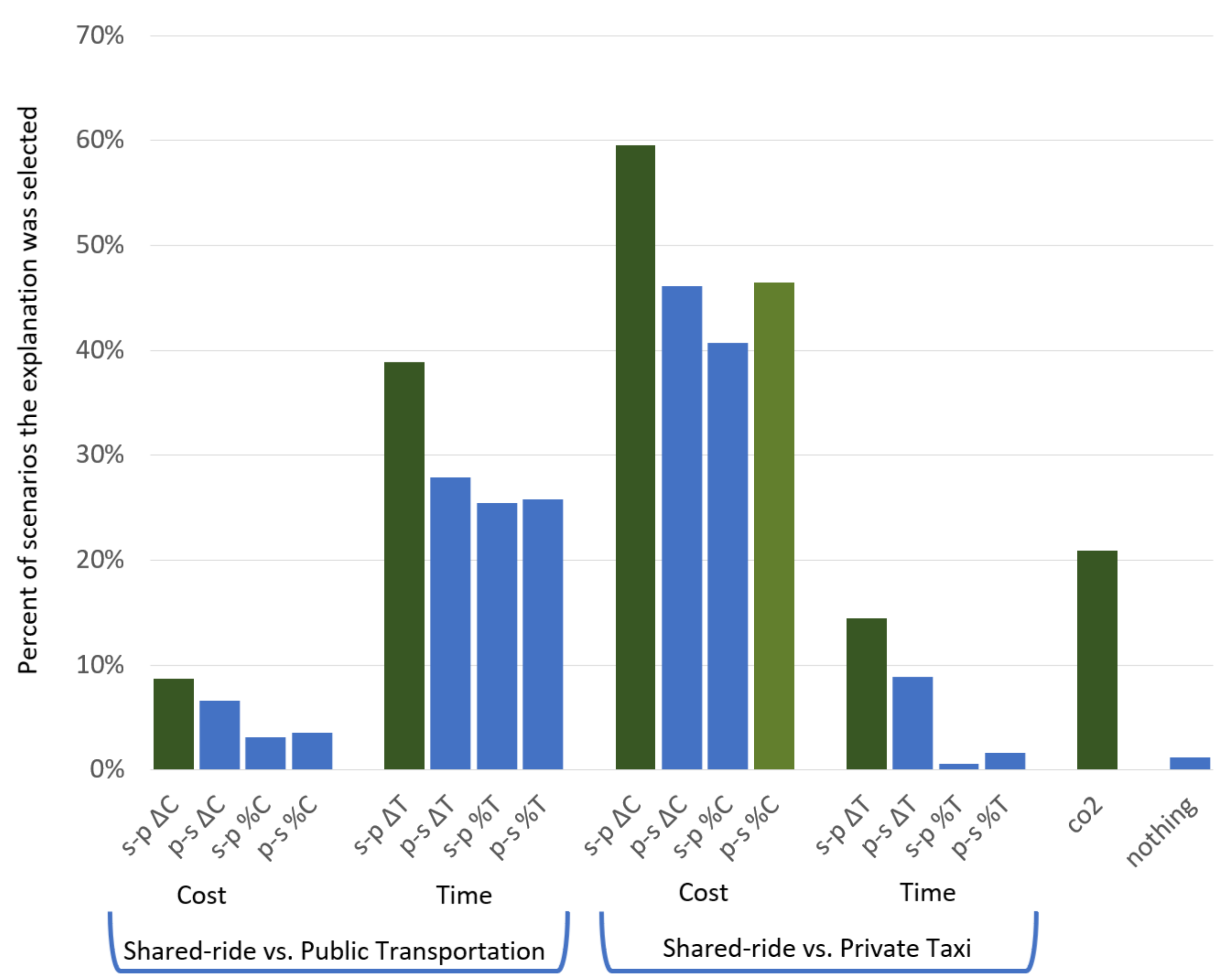}  
		\caption{The percent of scenarios that every explanation was selected in the first experiment.
			The explanations marked in green were selected for the second experiment.}
		\label{fig:humanResults1}
	\end{figure}
	
	\section{Results}
	
	Recall that the first experiment was designed to select the most appropriate explanations (out of the initial $17$ possible explanations). The results of this experiment are depicted in Figure~\ref{fig:humanResults1}. The x-axis describes the possible explanations according to the $4$ classes. Specifically, the factor from the anchoring class is denoted by s-p or p-s; s-p means that the explanation is from the shared-ride perspective, while p-s means that it is from the alternative (private/public) mode of transportation. The factor from the comparison criterion class is denoted by $\Delta$ or $\%$; $\Delta$ means that the explanation presents an absolute difference while $\%$ means that a relative difference is presented. We chose $6$ explanations for the next experiment, which are marked in green.

	\begin{figure}[hbpt]
		\centering
		\includegraphics[width=0.87\columnwidth]{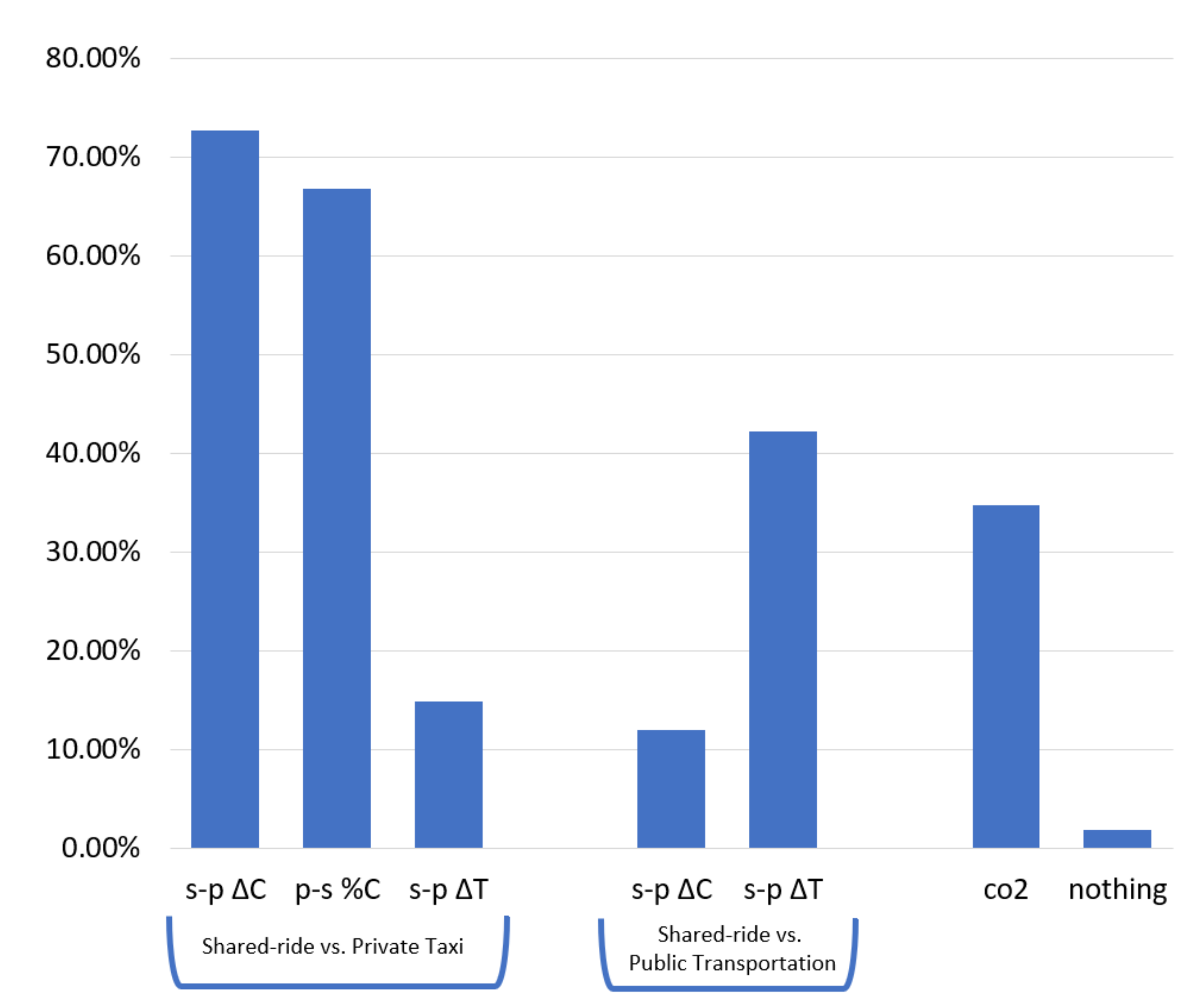} 
		\caption{The percent of scenarios that every explanation was selected in the second experiment. The obtained data-set was used to train AXIS.}
		\label{fig:humanResults2}
	\end{figure}
	
	As depicted by Figure \ref{fig:humanResults1}, the subjects chose explanations that compare the ride with a private taxi more often than those comparing the ride with public transportation. We believe that this is because from a human perspective a shared-ride resembles a private taxi more than public transportation.
	Furthermore, when comparing with a private taxi, the subjects preferred to compare the shared-ride with the \emph{cost} of a private taxi, while when comparing to public transportation, the subjects preferred to compare it with the travel time. This is expected, since the travel time by a private taxi is less than the travel time by a shared ride, so comparing the travel time to a private taxi is less likely to increase user satisfaction. 
	We also notice that with absolute difference the subjects preferred the shared ride perspective, while with relative difference the subjects preferred the alternative mode of transportation perspective. We conjecture that this is due to the higher percentages when using the alternative mode prospective. For example, if the shared ride saves $20\%$ of the cost when compared to a private ride, the subjects preferred the explanation that a private ride costs $25\%$ more.

	The second experiment was designed to collect data from humans on the most appropriate explanations (out of the $6$ chosen explanations) for each scenario. The results are depicted in Figure~\ref{fig:humanResults2}. This data was used to train AXIS. 
	The accuracy of the neural network on the test-set is $74.9\%$. That is, the model correctly predicts whether to provide a given explanation in a given scenario in almost $75\%$ of the cases.
	
\begin{figure}[hbpt]
		\centering
		\includegraphics[width=0.55\columnwidth]{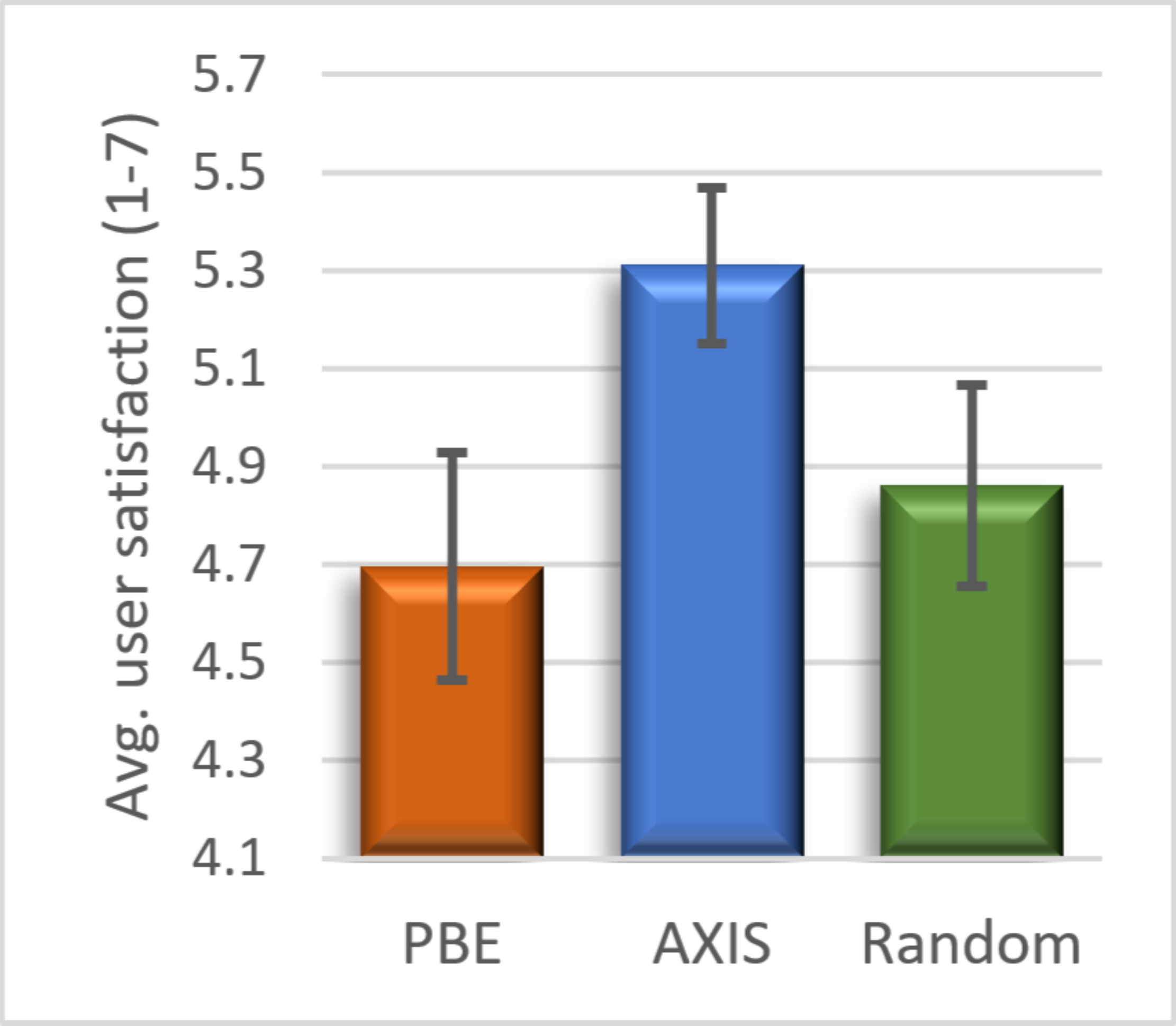} 
		\caption{A comparison between the performance of AXIS, the PBE agent and the random agent. The bars indicate the $95\%$ confidence interval.
			AXIS significantly outperformed both baseline agents $(p<0.001)$.}
		\label{fig:humanResults3}
	\end{figure}

	The third experiment was designed to evaluate AXIS against the PBE  agent; the results are depicted in Figure~\ref{fig:humanResults3}. 
	AXIS outperforms the PBE agent; the difference is statistically significant $(p<10^{-5})$, using the student t-test. 
	We note that achieving such a difference is non-trivial since the ride scenarios are identical and only differ by the information that is provided to the user.
	
	The forth experiment was designed to evaluate AXIS against the random baseline agent; the results are depicted in Figure~\ref{fig:humanResults3}. 
	AXIS outperforms the random agent; the difference is statistically significant $(p<0.001)$, using the student t-test. 
	We note that AXIS and the random agent provided a similar number of explanations on average ($2.551$ and $2.51$, respectively). That is, AXIS performed well not because of the number of explanations it provided, but since it provided appropriate explanations for the given scenarios. 
	
	%
	
	We conclude this section by showing an example of a ride scenario presented to some of the subjects, along with the information provided by the PBE agent, and the explanations selected by the random agent and by AXIS.
	In this scenario the subject is assumed to travel by a shared ride from JFK Station to 102-3 188th St, Jamaica, NY.
	The shared ride took $13$ minutes and cost $\$7.53$.
	The PBE agent provided the following information:
	\begin{itemize}
		\item ``A private ride would have cost $\$13.83$ and would have taken $12$ minutes''.
		\item ``Public transportation costs \$2.5 and would have taken 26 minutes''.
	\end{itemize}
	The random agent provided the following explanations:
	\begin{itemize}
		\item ``A private taxi would have cost $\$6.3$ more''.
		\item ``A ride by public transportation would have saved you only $\$5.03$''.
	\end{itemize}
	Instead, AXIS selected the following explanations:
	\begin{itemize}
		\item ``The shared ride had saved you $\$6.3$ over a private taxi''.
		\item ``A private taxi would have cost $83\%$ more''.
		\item ``The shared ride saved you $4$ minutes over public transportation''.
	\end{itemize}
	Clearly, the explanations provided by AXIS seem much more compelling.

	\section{Conclusions and Future Work}
	
	In this paper we took a first step towards the development of agents that provide explanations in a multi-agent system with a goal of increasing user satisfaction.
	We first modeled the explanation selection problem as a signaling game and determined the unique set of Perfect Bayesian Equilibria (PBE). We then presented AXIS, an agent that, when given a shared-ride along with its possible alternatives, selects the explanations that are most likely to increase user satisfaction.
	We ran four experiments with humans. The first experiment was used to narrow the set of possible explanations, the second experiment collected data for the neural network to train on, the third experiment was used to evaluate the performance of AXIS against that of the PBE agent, and the fourth experiment was used to evaluate the performance of AXIS against that of an agent that randomly chooses explanations. We showed that AXIS outperforms the other agents in terms of user satisfaction.
	
	In future work we will consider natural language generation methods for generating explanations that are likely to increase user satisfaction. We also plan to extend the set of possible explanations, and to implement user modeling in order to provide explanations that are appropriate not only for a given scenario but also for a given specific user.
	

	\section*{Acknowledgment}
	This research was supported in part by the Ministry of Science, Technology \& Space, Israel.
	
	\bibliographystyle{abbrv}
	\bibliography{main}
	
	\section*{Appendix}
	\subsection*{Attention Check Question}
	Which of the following claims do you agree with?
	\begin{itemize}
		\item A shared ride may take longer than a private ride.
		\item A shared ride is supposed to be more expensive than a private ride.
		\item The cost of public transportation is usually less than the cost of a private ride.
		\item In a private ride there are at least 3 passengers.
		\item Public transportation usually takes longer than a  private ride.
	\end{itemize}
	
	\subsection*{The Text for the First Two Experiments}
	In this survey we would like to learn which explanations increase the satisfaction from a ride sharing service.
	Suppose that you are a representative of a ride sharing service. This service assigns multiple passengers with different destinations to a shared taxi and divides the cost among them. Assume that the customer of your service has just completed the shared ride.
	Below are given few scenarios for a shared ride. For each of the scenarios you should choose one or more suitable explanation(s) that you believe will increase the customer's satisfaction.
	
	\subsection*{The Text for the Third and Fourth Experiments}
	In this survey we would like to evaluate your satisfaction from using shared taxi services.
	The following questions contain description of some shared rides. Please provide your rate of satisfaction from each ride on a scale from 1 to 7.
	Please read the details carefully and try to evaluate your satisfaction in each scenario as accurate as possible. Good luck!

\end{document}